\documentclass[11pt,a4paper]{article}
\usepackage[hyperref]{emnlp-ijcnlp-2019}
\usepackage{times}

\usepackage{latexsym}
\usepackage{graphicx}
\usepackage{amsmath}
\usepackage{amssymb}
\usepackage{algorithm}% http://ctan.org/pkg/algorithm

\usepackage{array} 
\usepackage{booktabs} 
\usepackage{multirow}

\usepackage{url}

\usepackage{subcaption}

\newcommand{\argmax}{\operatornamewithlimits{argmax}}

\aclfinalcopy % Uncomment this line for the final submission

\usepackage{makecell}

\newcommand{\insertEnsembleDiagramsSubcaption}{
\begin{figure}[!t]
\centering
\begin{subfigure}{.5\columnwidth}
  \centering
  \includegraphics[width=.90\linewidth]{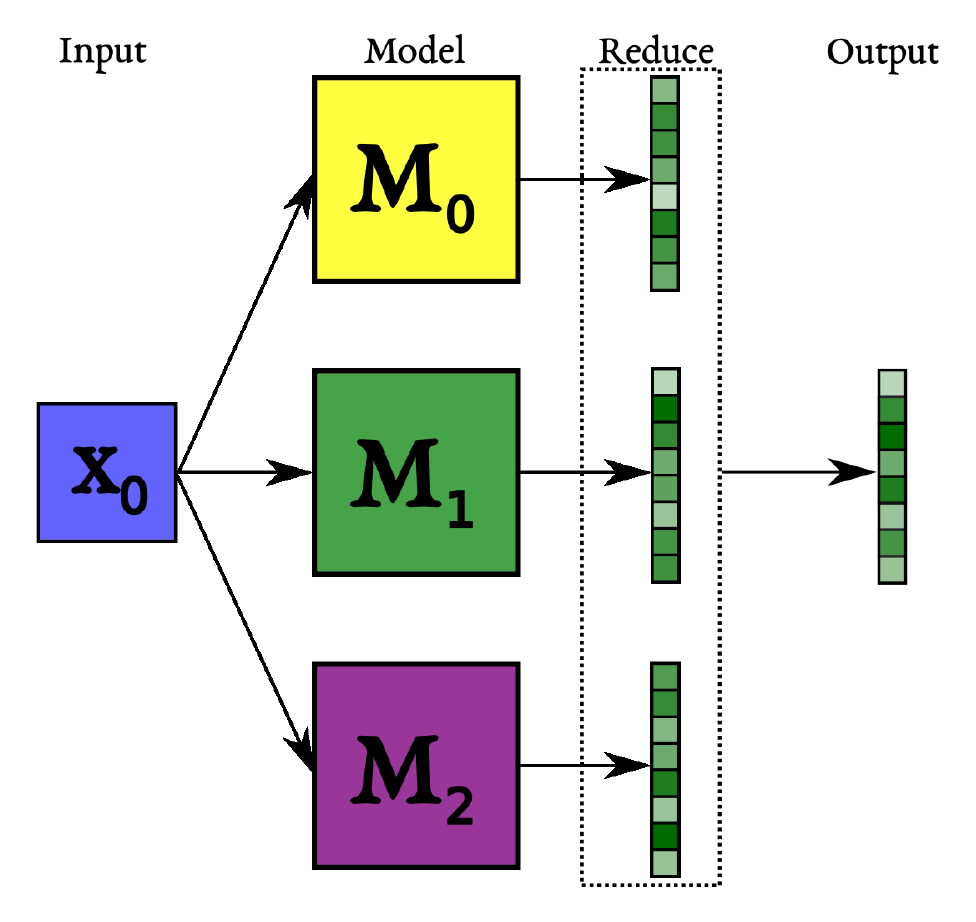}
  \caption{}
  \label{fig:sub1}
\end{subfigure}%
\begin{subfigure}{.5\columnwidth}
  \centering
  \includegraphics[width=.90\linewidth]{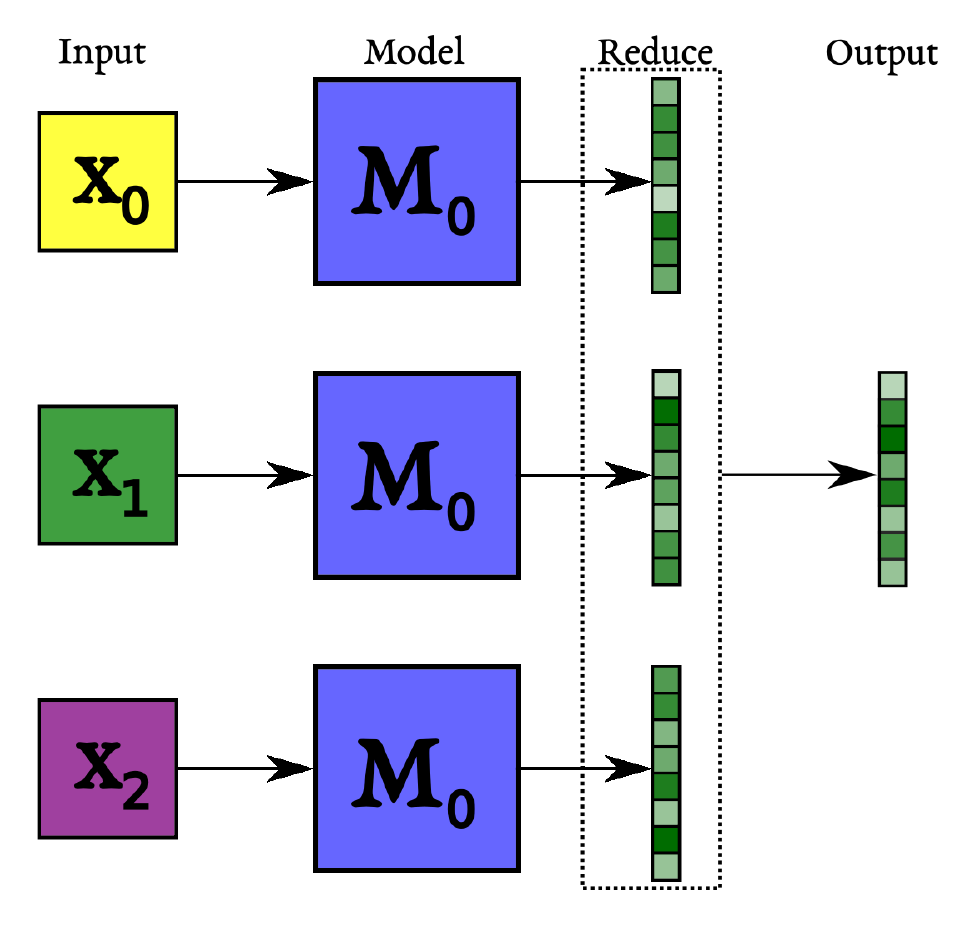}
  \caption{}
  \label{fig:sub2}
\end{subfigure}
\caption{ (a) In standard ensembling, a single input is passed through distinct models. (b) In dynamic ensembling distinct, complimentary inputs are passed through the same model. In both cases, the set of outputs is combined via the \texttt{reduce} operation.}
\label{fig:dynamic_ensemble_diagram}
\end{figure}
}

\newcommand{\insertWCEPEvaluationTable}{

\begin{table}[]
\centering
\small
\begin{tabular}{l|lll}
\hline
 \multicolumn{4}{c}{\textbf{F-score}} \\
 Method & R1 & R2 & RL \\ 
 \hline
\textsc{Oracle (Multi)} & 0.558 & 0.29 & 0.4  \\
\textsc{Oracle (Single)} & 0.539 & 0.283 & 0.401  \\
\textsc{Lead Oracle} & 0.329 & 0.131 & 0.233  \\
\textsc{Random Lead} & 0.276 & 0.091 & 0.206  \\ 
\textsc{Random} & 0.181 & 0.03 & 0.128  \\ 
\hline
\textsc{TextRank}       & 0.341 & 0.131 & 0.25  \\
\textsc{Centroid}       & 0.341 & 0.133 & 0.251  \\
\textsc{Submodular}     & 0.344 & 0.131 & 0.25  \\
\textsc{TSR}            & 0.353 & 0.137 & 0.257  \\
\textsc{BertReg}        & 0.35 & 0.135 & 0.255 \\
\textsc{Submodular+Abs} & 0.306 & 0.101 & 0.214  \\ 
\hline
\textsc{bart-cnn-dm DynE-1}  & 0.27  & 0.083 & 0.201 \\ 
\textsc{bart-cnn-dm DynE-5}  & 0.303 & 0.097 & 0.223 \\ 
\hline
\textsc{bart-wcep DynE1}  & 0.328 & 0.13  & 0.237 \\ 
\textsc{bart-wcep DynE-5} & 0.354  & 0.151  & 0.256  \\ 
\hline

\end{tabular}
\caption{Evaluation results on WCEP test set.}
\label{tab:wcep-evaluation}
\end{table}

}

\newcommand{\insertDecodingHyperparametersTable}{
\begin{table}[]
\centering
\small
\begin{tabular}{l|lll}
\hline
 Dataset & \textsc{src-len} & \textsc{tgt-len} & \textsc{num-beams} \\ 
 \hline
\textsc{WCEP}  & 512  & 64 & 5 \\ 
\textsc{MultiNews}  & 768  & 256 & 3 \\ 
\textsc{MultiNews}  & 768  & 128 & 5 \\ 
\hline
\end{tabular}
\caption{Decoding hyperparameters for each MDS dataset.}
\label{tab:dataset-decoding-hyperparameters}
\end{table}

}

\newcommand{\insertMultiNewsEvaluationTable}{

\begin{table}[]
\centering
\small
\begin{tabular}{l|lll}
\hline
 \multicolumn{4}{c}{\textbf{F-score}} \\
 Method & R1 & R2 & RL \\ 
 \hline
\textsc{MultiNews Transformer}  & 0.42  & 0.149  & 0.195 \\ 
\textsc{MultiNews Hi-Map}  & 0.408  & 0.149  & 0.197 \\ 
\hline
\textsc{bart-MultiNews DynE-1}  & 0.439 & 0.158  & 0.222 \\ 
\textsc{bart-MultiNews DynE-5} & 0.432 & 0.136  & 0.204 \\ 
\hline

\end{tabular}
\caption{Evaluation results on MultiNews test set.}
\label{tab:multinews-evaluation}
\end{table}

}

\newcommand{\insertDUCEvaluationTable}{

\begin{table}[]
\centering
\small
\begin{tabular}{l|lll}
\hline
\multicolumn{4}{c}{\textbf{F-score}} \\
Method & R1 & R2 & R-SU \\ 
\hline
\textsc{PG-MMR w/ SummRec}          & 34.57 & 7.46 & 11.36 \\
\textsc{PG-MMR w/ SentAttn}         & 36.52 & 8.52 & 12.57 \\
\textsc{PG-MMR w/ Cosine (default)} & 36.88 & 8.73 & 12.64 \\
\hline 
\textsc{bart-cnn-dm DynE-1} & 25.95 & 5.41  & 8.22 \\ 
\textsc{bart-cnn-dm DynE-5} & 32.64 & 7.78  & 11.22 \\ 
\textsc{bart-cnn-dm DynE-8} & 33.21 & 8.06  & 11.47 \\ 

\hline

\end{tabular}
\caption{Evaluation results on DUC2004 MDS test set, compared with Lebanoff et al 2018.}
\label{tab:duc2004-evaluation}
\end{table}

}

\newcommand{\insertDecodingHeatmap}{
\begin{figure*}[t]
\centering
  \centering
  \includegraphics[width=.80\linewidth]{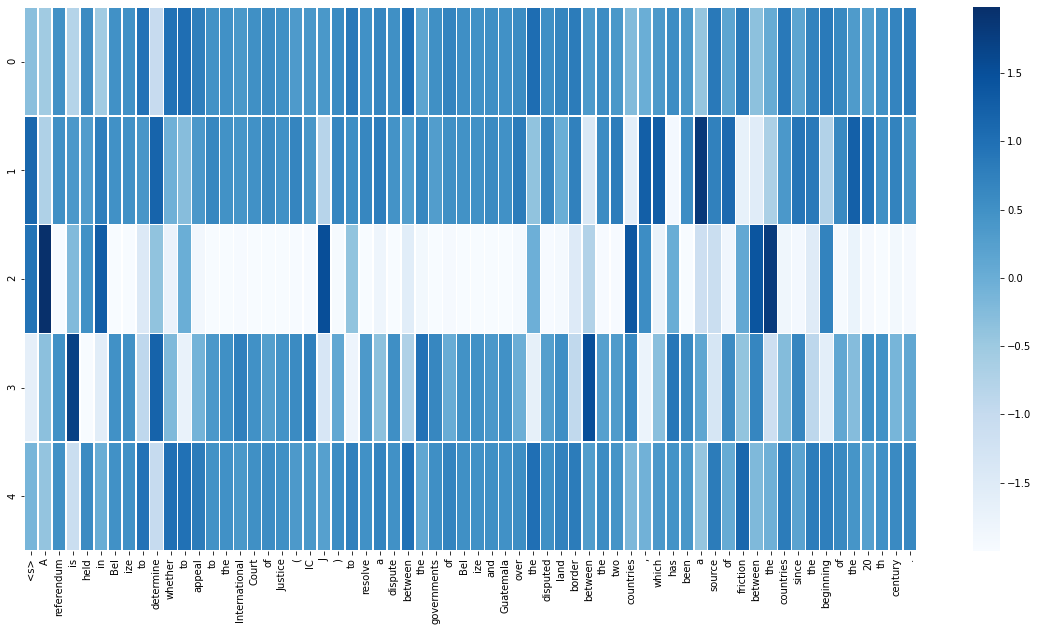}
  \caption{Visualizing ensemble scores per-input, per-timestep on an example from the WCEP MDS dataset. The example at index 2 is deliberately selected from another cluster, and thus does not align with the content of other cluster items. The visualization convincingly shows that dynamic ensembling effectively deals with this noise article in the cluster.}
  \label{fig:decoding-heatmap}
\end{figure*}
}

\title{DynE: Dynamic Ensemble Decoding for Multi-Document Summarization}

\author{Chris Hokamp$^1$, Demian Gholipour Ghalandari$^{1,2}$, Nghia The Pham$^1$, John Glover$^1$ \\
$^1$Aylien Ltd., Dublin, Ireland \\
$^2$Insight Centre for Data Analytics, University College Dublin, Ireland \\
$^1$\texttt{\{first-name\}@aylien.com}}

\date{}

\begin{document}
\maketitle
\begin{abstract}
   Sequence-to-sequence (s2s) models are the basis for extensive work in natural language processing. However, some applications, such as multi-document summarization, multi-modal machine translation, and the automatic post-editing of machine translation, require mapping a set of multiple distinct inputs into a single output sequence.  Recent work has introduced bespoke architectures for these multi-input settings, and developed models which can handle increasingly longer inputs; however, the performance of special model architectures is limited by the available in-domain training data. In this work we propose a simple decoding methodology which ensembles the output of multiple instances of the same model on different inputs. Our proposed approach allows models trained for vanilla s2s tasks to be directly used in multi-input settings. This works particularly well when each of the inputs has significant overlap with the others, as when compressing a cluster of news articles about the same event into a single coherent summary, and we obtain state-of-the-art results on several multi-document summarization datasets. 
\end{abstract}

\section{Introduction}
The practice of pre-training large neural networks using self-supervision and then fine-tuning on downstream tasks has produced new state-of-the-art results in many application areas.
Although originally used for classification and sequence labeling, these models have recently been adopted for sequence-to-sequence problems as well \cite{lample2019cross,lewis2019bart,2019t5}.
There is also been renewed interest in the task of multi-document summarization (MDS), highlighted by the introduction of several new large-scale datasets \cite{multinews2019,newshead2020,ghalandari2020WCEP}.

\insertEnsembleDiagramsSubcaption

% Building on promising results in single-document summarization (SDS), state-of-the-art results on MDS have been achieved by re-pu
% MDS can be naturally framed as compression or summarization of an unordered set of complimentary inputs. Models which map more than one sequence to a single sequence are a natural extension of the sequence-to-sequence (\textbf{s2s}) setting. In the rest of this paper, we will use the abbreviations \textbf{s2s} and \textbf{m2s} to distinguish between these common setups.\john{should we define what m2s stands for explicitly here? Also seems a bit close to MDS, maybe ms2s is clearer?}

% \john{Slight reframe as I'm not sure if we can really make a claim about the general case here if we only evaluate on MDS}
In this work we explore the use of powerful pre-trained models on the MDS task, which we view as an instance of the more general multi-input sequence-to-sequence problem. These models are data-hungry, and expensive to train, thus we would like to establish to what extent models trained on single inputs can be directly used for multi-input tasks. For applications such as MDS, model architectures may be adapted to better suit the specific task, potentially improving performance at the cost of limiting the generality of the method. An alternative approach is to simply change the way predictions are generated from a single-document summarization (SDS) model, enabling the reuse of existing SDS models for MDS \cite{lebanoff-etal-2018-adapting}.

Ensembling is a general technique that is used extensively to improve the performance of machine learning models. In most cases, ensembling means combining the outputs of distinct models on the same input. However, for applications where multiple inputs are available, we may also consider ensembles which combine the outputs of the same model on different inputs (see figure \ref{fig:dynamic_ensemble_diagram}). This is especially appealing in cases where the inputs are complimentary or contain significantly overlapping information, and we wish to leverage consensus in the inputs to make better predictions about what content is important. In settings where the inputs are expected to be highly overlapping in content, or paraphrases of the same content, we expect that simply ensembling model output over each of these inputs should improve performance, and we verify this hypothesis experimentally in \autoref{sec:experiments}. 

We show that pre-trained encoder-decoder models fine-tuned for single document abstractive summarization can achieve state-of-the-art performance on the multi-document summarization task without any changes to the SDS model architecture or training objectives. The key contributions of this work are:

\begin{enumerate}
    \item A simple ensembling method is presented that allows models trained for single inputs to be used in multi-input settings. 
    \item We show that our method achieves state-of-the-art performance on multi-document summarization tasks with no changes to the underlying model architecture.
    \item We show how dynamic ensembling also enables some degree of interpretability and provenance tracing by collecting statistics during the decoding process.
\end{enumerate}

% a difference between Lebanoff and us is that we can completely treat model as black box. 
% ensembling intuition is that models have complimentary information -- we apply this same intuition to the input document cluster for MDS.
% complexity of interpretation increases as number of inputs increase

% Recent work has explored extensions to SDS to enable MDS. However, these models are purpose-built for the MDS task, which limits their generality. 
% comment on efficiency and simplicity -- no task-specific pretraining required.

% define ensembling

% typically ensembling means combining the output of different models with the same input, here 
% \textit{dynamic ensembling} means combining the output of the same model with different inputs.

% Motivate: multi-document training datasets may be very limited in many domains, but we may still wish to leverage redundant information from many documents when such information is present.

\section{Background}

\subsection{Multi-Document Summarization Datasets}

Traditional datasets for the MDS task include \textsc{DUC 2004}  \cite{paul2004introduction} and \textsc{TAC 2011} \cite{owczarzak2011overview}.  Recently, the \textsc{MultiNews} dataset \cite{multinews2019} was introduced, containing 56,000 clusters with an average of 2.3 source documents per cluster. 

\citet{ghalandari2020WCEP} presented the \textsc{WCEP} dataset, which is a large-scale collection of clusters of news articles with a corresponding summary, constructed using the Wikipedia Current Events Portal, with additional articles gathered from CommonCrawl. This dataset is intended to mimic real-world MDS settings, with a large number of documents per-cluster, a large number of clusters, and acknowledgement of some degree of noise in the clusters. The \textsc{NewSHead} dataset \citep{newshead2020} is a recently introduced large-scale MDS dataset for headline generation. 

\citet{zopf2018auto} created the \textsc{auto-\textit{h}MDS} dataset by using the lead section of Wikipedia articles as summaries, and automatically searching for related documents on the web, resulting in 7,300 clusters. The \textsc{WikiSum} dataset \cite{liu2018generating} uses a similar approach and additionally uses cited sources on Wikipedia. The dataset contains 2.3 million clusters. These Wikipedia-based datasets have long summaries about various topics and heterogeneous source documents, whereas our focus is on short summaries of highly similar news articles.

The summaries provided by different MDS datasets are diverse in terms of summary length. \textsc{WCEP} summaries are one or two sentences, resembling headlines or short TL;DR summaries, while \textsc{MultiNews} summaries are often longer than some of the input articles themselves, and often look like standalone news stories.

\subsection{Multi-Document Summarization Models}

\paragraph{Extractive Models}
Most MDS approaches to date are based on sentence extraction \cite{goldstein2000multi, radev2004centroid, gillick2009scalable, haghighi2009exploring, lin2011class, hong2014improving, yasunaga2017graph}. 
Many of these approaches are very efficient due to sparse text representations and simple greedy selection algorithms \cite{zopf2018scores}. Other approaches use explicit sentence compression and fusion \cite{ ganesan2010opinosis, berg2011jointly, banerjee2015multi, chali2017towards, nayeem2018abstractive}, which may be  considered a hybrid of extractive and abstractive summarization techniques.

\paragraph{Abstractive Models}
\citet{lebanoff-etal-2018-adapting} emphasize the important practical need to adapt models trained for SDS to the MDS task. \citet{lebanoff-etal-2018-adapting} allow SDS models to be used for MDS by modifying the attention weights. Like ours, their approach does not require training on MDS datasets, but it is specific to certain model  architectures, and it is unclear how generalizable this approach may be. Their approach to avoiding redundancy by dynamically modifying attention weights is complementary to the dynamic ensembling method we present.  

\citet{LiuL19-hierarchical,lebanoff-etal-2018-adapting,multinews2019,ElSahar2020SelfSupervisedAC} create bespoke model architectures for MDS. However, the MDS task as currently defined focuses upon utilizing the overlap of content in multiple documents to improve summaries. Therefore, we hypothesize that special architectures may be unnecessary for this task in its current formulation. 

\paragraph{Pre--trained Text Generation Models}

Large pre-trained transformer-based models which can be efficiently fine-tuned for downstream tasks are the state-of-the-art for many classification and sequence labeling tasks \cite{devlin2019bert,liu2019roberta,clark2019electra}, inter alia. Recent work has shown that pre-trained models can be fine-tuned to achieve strong performance on sequence-to-sequence problems as well, by framing the pre-training task as some variant of denoising auto-encoding. When fine-tuned for the SDS task, these models achieve state-of-the-art performance \cite{lewis2019bart,2019t5}. \citet{lewis2019bart} introduced the BART model, which we use to validate the ideas in this work. 

% \subsection{Multi-input Models}

% \chris{mention both extractive and abstractive designs, extractive has a big efficiency advantage, esp as clusters become very large}

% Many real world applications of machine learning are naturally viewed as multi-input settings, where each distinct input potentially includes additional information that can help inform prediction. Notably, bespoke models multi-modal MT and multi-source MT have recently been investigated.

% extend background work beyond summarization -- MDS is just the testbed, there are many other possible application areas for this 
% grammatical error correction, multi-input MT, MDS, ..., splitting inputs into pieces, etc... 

\section{Model}

We consider a dataset $ \mathcal{D} $ consisting of pairs $ \mathcal{X}, \mathbf{y} \in \mathcal{D}$, where each $ \mathcal{X} $ consists of one or more distinct sequences $ \mathbf{x} \in \mathcal{X} $, and $ \mathbf{y} $ is a single output sequence. Given a set of documents $ \mathcal{X} $, the model should predict the best output summary $ \mathbf{\hat{y}^{*}} $. In the case of MDS, we refer to each $ \mathcal{X} $ as a \textit{cluster} since it consists of one or more documents that should be about the same event or topic. 

Considering a set of distinct inputs $ \mathcal{X} $, each of which are fed to the model independently, we can jointly decode by ensembling the predictions of an SDS model given each distinct input. At each decoding timestep $ t $, the output of the model with respect to each input is computed, then a \textit{reduce} function combines the individual outputs into a single output. Note that the prefix $ \mathbf{y}_{0:t-1} $ is the same for all inputs; thus, the predictions of all model instances are conditioned on the same shared partial output sequence constructed so far. The average log-probability assigned to a particular token $y$ at decoding timestep $ t $ is (Eq.~\ref{eq:ensemble_best_y_t}):

% ensemble probability of a single vocabulary item
\begin{equation}
p_{\theta}(y_{t} | \mathcal{X}) = 
    \frac{1}{|\mathcal{X}|} 
    \sum\limits_{\mathbf{x_{i}} \in \mathcal{X}}
    p_{\theta}(y_{t} | \mathbf{x}_{i}; \mathbf{y}_{0:t-1}),
\label{eq:ensemble_best_y_t}
\end{equation}

and the global score of a finished sequence $ \mathbf{\hat{y}}$ according to the ensemble is (Eq.~\ref{eq:ensemble_best_y}):

\begin{equation}
p_{\theta}(\mathbf{\hat{y}} | \mathcal{X}) = 
    \sum\limits_{t=0}^{T} 
    p_{\theta}(y_{t} | \mathcal{X})
\label{eq:ensemble_best_y}
\end{equation}

We use standard beam-search decoding to search for the best $ \mathbf{\hat{y}^{*}} $:

% First define word-level (timestep-level loss), then accumulate over seqs and minibatches to get full training loss
\begin{equation}
\mathbf{\hat{y}^{*}} = \argmax\limits_{\mathbf{y} \in \{\mathbf{y^{[T]}}\}} p_{\theta}(\mathbf{y} | \mathcal{X}),
\label{eq:find_best_y}
\end{equation} 

\noindent where $ \{\mathbf{y^{[T]}}\} $ is the set of all possible sequences up to a maximum length $ \mathbf{T} $.

We refer to this approach as Dynamic Ensembling (\textsc{DynE}). We refer to the function which combines model output conditioned on each input into a single output as the \textit{reduce} function. In this case, we simply take the elementwise mean of output probabilities, but in general arbitrary reduce functions are possible that would weight or combine the contribution of each input differently based upon heuristics or auxiliary models\footnote{Note in principle it also is possible to ensemble both distinct models and distinct inputs simultaneously}.

% models take the form f(X) --> y, which we factorize over the individual sequences in X
% TODO: (k)-argmax equation (short version)

% TODO: Explain bidir encoder-->denoising autoencoder as pretrained architecture
%  for seq2seq tasks

% probability of sequence
% \begin{equation}
% \mathbf{\hat{y}} = \argmax\limits_{\mathbf{y} \in \{\mathbf{y^{[T]}}\}} p_{\theta}(\mathbf{y} | \mathbf{x}),
% \label{eq:find_best_y}
% \end{equation}

% timestep factorized probability of sequence
% \begin{equation}
% p_{\theta}(\mathbf{y} | \mathbf{x}) = \prod_{t=0}^{T} p_{\theta}(y_{t} | \mathbf{x}; \{ y_{0} \ldots y_{t-1}\}).
% \label{eq:factorize_best_y}
% \end{equation}

% model ensembled timestep factorized probability of sequence
%Considering a set of distinct models $ \mathbf{M} $, one way to combine the outputs of each of these models together in an ensemble is: 

% If we change our view of an ensemble to an ensemble over a set of distinct \emph{inputs}, (EQUATION X) becomes: 

\section{Experiments}
\label{sec:experiments}

For all experiments, we start from the pre-trained \texttt{bart-large} model fine-tuned on the \texttt{cnn-dm} summarization dataset\footnote{model id: \texttt{bart-large-cnn}}. This model achieves state-of-the-art performance on the \texttt{cnn-dm} SDS dataset \cite{nallapati-etal-2016-abstractive}, so we hypothesize that it is a good base model for MDS as well. Our implementation is based on the \texttt{transformers} library \cite{Wolf2019HuggingFacesTS}\footnote{code, outputs, and trained model checkpoints available at: \url{https://github.com/chrishokamp/dynamic-transformer-ensembles}}.

We evaluate \textsc{DynE} on three multi-document summarization datasets: \textsc{DUC2004} \citep{paul2004introduction}, \textsc{MultiNews} \citep{multinews2019}, and \textsc{WCEP} \citep{ghalandari2020WCEP}, and compare with existing MDS approaches.

% TODO: WCEP -- we actually only do baseline on WCEP
% For each dataset, we first evaluate the baseline performance of the BART model tuned on \texttt{cnn-dm} using DynE input clusters of one vs. five randomly selected documents. For MultiNews and WCEP, we then measure the improvement gained from fine-tuning. 

To fine-tune the SDS model on MDS datasets, we use the first document from a cluster as the input, and try to predict the reference summary. Note that the fine-tuning process is only a means of domain adaptation because the model is never aware of the multi-document  setting that we use during evaluation. For both \textsc{MultiNews} and \textsc{WCEP}, we notice a significant improvement after fine-tuning the summarization model for each domain. We select the best checkpoint according to performance on the development set, and use this checkpoint to evaluate performance on the test set. 

For \textsc{WCEP} and \textsc{MultiNews}, we present results using the same evaluation as \citet{ghalandari2020WCEP}, while for \textsc{DUC2004} we evaluate with the same script as \citet{lebanoff-etal-2018-adapting} for fair comparison. Because each dataset has a different average summary length and a different average length of input articles, we use different decoding hyper-parameters for each dataset at inference time (Table \ref{tab:dataset-decoding-hyperparameters}).

% Because there are significant differences between the evaluations in \citet{lebanoff-etal-2018-adapting}, \citet{multinews2019}, and \citet{ghalandari2020WCEP}, we present full results with all three evaluation workflows in the appendix, and we note that system rankings do not change accross the different evaluation workflows\footnote{All outputs and evaluation code are available in the repository for this work}.

% Note in many cases clusters have less than 5 articles, in such cases we select everything

\insertDecodingHyperparametersTable

\subsection{WCEP} 

\insertWCEPEvaluationTable

Table \ref{tab:wcep-evaluation} presents scores on the \textsc{WCEP} dataset, compared with oracle performance and with several strong extractive models. Our ensembling method excels on this dataset, likely because it is highly abstractive, with short reference summaries. Fine-tuning the base model leads to significant improvement, and adding more documents to input clusters consistently increases summary quality. An ensemble with five documents per input cluster outperforms all extractive baselines. Note that the truncated version of \textsc{WCEP} contains clusters of up to 100 documents, but we use only up to five randomly selected documents from each cluster. Thus the performance ceiling for abstractive MDS on this dataset is potentially much higher.

\subsection{MultiNews} 

\insertMultiNewsEvaluationTable

Table \ref{tab:multinews-evaluation} shows results on the \textsc{MultiNews} dataset, using the output of the two best models from \citet{multinews2019} as baselines.  Using only one randomly selected document already outperforms the state-of-the-art on this dataset, but adding more inputs into the ensemble hurts performance slightly. This is an unexpected result which warrants further investigation; however, preliminary inspection reveals that some clusters in \textsc{MultiNews} contain a document which is very similar to the reference summary, thus in the case that this particular document is selected as input, the evaluation results are artificially high. We plan to investigate issues with the \textsc{MultiNews} dataset further when time allows.

% (see \autoref{sec:analysis}).

\subsection{DUC 2004} 

\insertDUCEvaluationTable

Table \ref{tab:duc2004-evaluation} contains the results of \textsc{DynE} with clusters of various sizes on the \textsc{DUC2004} dataset. For this dataset, we cannot fine tune because we do not have a held-out training dataset. Despite this, performance improves as we add more documents to input clusters, and approaches the results of strong abstractive baselines.

% TODO: expand or delete?
% \subsection{Reduce Functions}

% For a given input cluster, weighting each document's contributions using information from the other documents is likely to improve performance. 

% Scoring each document using the other documents in the cluster is a promising means of weighting each input's contribution beyond a simple mean. We study how outputs change when we heuristically remove documents that do not agree with other documents in the ensemble. 
% END: expand or delete?

\section{Analysis}
\label{sec:analysis}
\insertDecodingHeatmap

\subsection{Interpretibility of Multi-Document Summaries}

An advantage of inference-time ensembling is that the timestep-level scores are simply composed from predictions on the individual inputs. Utilizing the global ensemble predictions, but conditioning on each distinct input at each timestep, we can directly study how each input document contributed to the ensemble output at that timestep. By combining the timestep-level scores, we can also measure how likely the final output is according to each input. Unlike attention-based interpretations, or other model-specific approaches to interpretability, this method is exact in the sense that the contribution of each input does not depend upon the other inputs.

The score that each input $ \mathbf{x}_{i} $ assigns to a token at each decoding timestep is (Eq.~\ref{eq:single_input_best_y_t}):

\begin{equation}
p_{\theta}(y_{t} | \mathbf{x}_i) = 
    p_{\theta}(y_{t} | \mathbf{x}_{i}; \mathbf{y}_{0:t-1}),
\label{eq:single_input_best_y_t}
\end{equation}

\noindent noting again that the prefix $ \mathbf{y}_{0:t-1} $ is the same for all inputs since we are decoding jointly. By collecting the scores for each input, we can visualize the contribution of each input to the ensemble's prediction (figure \ref{fig:decoding-heatmap}).

\section{Conclusion}

We have presented \textsc{DynE}, a simple approach to ensembling for multi-input problems, allowing SDS models to be directly leveraged to achieve state-of-the-art results on the MDS task. The ensembling approach presented here is well-suited to problems like multi-document summarization, where the information overlap between individual inputs is presumed to be high; however, for problems where the inputs contain disjoint information, a more sophisticated decoding controller with global awareness of all inputs may be needed.  

A weakness of the method proposed here is that document length is not a factor in the relative contribution of input documents to summaries. For news data in particular, it is plausible that a short document such as a tweet could add relevant novel information to that provided by the longer documents in the cluster. Another weakness is that the decoder memory usage increases linearly with the number of inputs, despite the decoded prefix being the same for all inputs. However, we note that \textsc{DynE} allows parallel decoding of the documents in a cluster, in contrast to other methods which necessarily combine all documents in a cluster into a single input. In future work, we hope to address some of the weaknesses with \textsc{DynE} and apply the technique to other tasks and model types.

% disjoint would be i.e. when we split a single document into paragraphs and decode each one independently

% Other met hods also don't account for this issue
% \section*{Acknowledgments}
% 
% The acknowledgments should go immediately before the references.  Do
% not number the acknowledgments section. Do not include this section
% when submitting your paper for review. \\
% 
% \noindent {\bf Preparing References:} \\
% 
% Include your own bib file like this:
% {\small\verb|\bibliographystyle{acl_natbib}|
% \verb|\bibliography{emnlp-ijcnlp-2019}|}
% 
% Where \verb|emnlp-ijcnlp-2019| corresponds to the {\tt emnlp-ijcnlp-2019.bib} file.

\bibliography{mds}

\begin{thebibliography}{31}
\expandafter\ifx\csname natexlab\endcsname\relax\def\natexlab#1{#1}\fi

\bibitem[{Banerjee et~al.(2015)Banerjee, Mitra, and
  Sugiyama}]{banerjee2015multi}
Siddhartha Banerjee, Prasenjit Mitra, and Kazunari Sugiyama. 2015.
\newblock Multi-document abstractive summarization using ilp based
  multi-sentence compression.
\newblock In \emph{Proceedings of the 24th International Conference on
  Artificial Intelligence}, pages 1208--1214. AAAI Press.

\bibitem[{Berg-Kirkpatrick et~al.(2011)Berg-Kirkpatrick, Gillick, and
  Klein}]{berg2011jointly}
Taylor Berg-Kirkpatrick, Dan Gillick, and Dan Klein. 2011.
\newblock Jointly learning to extract and compress.
\newblock In \emph{Proceedings of the 49th Annual Meeting of the Association
  for Computational Linguistics: Human Language Technologies-Volume 1}, pages
  481--490. Association for Computational Linguistics.

\bibitem[{Chali et~al.(2017)Chali, Tanvee, and Nayeem}]{chali2017towards}
Yllias Chali, Moin Tanvee, and Mir~Tafseer Nayeem. 2017.
\newblock Towards abstractive multi-document summarization using submodular
  function-based framework, sentence compression and merging.
\newblock \emph{IJCNLP 2017}, page 418.

\bibitem[{Clark et~al.(2020)Clark, Luong, Le, and Manning}]{clark2019electra}
Kevin Clark, Minh-Thang Luong, Quoc~V. Le, and Christopher~D. Manning. 2020.
\newblock {ELECTRA}: Pre-training text encoders as discriminators rather than
  generators.
\newblock In \emph{ICLR}.

\bibitem[{Devlin et~al.(2019)Devlin, Chang, Lee, and
  Toutanova}]{devlin2019bert}
Jacob Devlin, Ming-Wei Chang, Kenton Lee, and Kristina Toutanova. 2019.
\newblock Bert: Pre-training of deep bidirectional transformers for language
  understanding.
\newblock In \emph{Proceedings of the 2019 Conference of the North American
  Chapter of the Association for Computational Linguistics: Human Language
  Technologies, Volume 1 (Long and Short Papers)}, pages 4171--4186.

\bibitem[{ElSahar et~al.(2020)ElSahar, Coavoux, Gall{\'e}, and
  Rozen}]{ElSahar2020SelfSupervisedAC}
Hady ElSahar, Maximin Coavoux, Matthias Gall{\'e}, and Jos Rozen. 2020.
\newblock Self-supervised and controlled multi-document opinion summarization.
\newblock \emph{ArXiv}, abs/2004.14754.

\bibitem[{Fabbri et~al.(2019)Fabbri, Li, She, Li, and Radev}]{multinews2019}
Alexander~Richard Fabbri, Irene Li, Tianwei She, Suyi Li, and Dragomir~R.
  Radev. 2019.
\newblock \href {https://www.aclweb.org/anthology/P19-1102/} {Multi-news: {A}
  large-scale multi-document summarization dataset and abstractive hierarchical
  model}.
\newblock In \emph{Proceedings of the 57th Conference of the Association for
  Computational Linguistics, {ACL} 2019, Florence, Italy, July 28- August 2,
  2019, Volume 1: Long Papers}, pages 1074--1084.

\bibitem[{Ganesan et~al.(2010)Ganesan, Zhai, and Han}]{ganesan2010opinosis}
Kavita Ganesan, ChengXiang Zhai, and Jiawei Han. 2010.
\newblock Opinosis: A graph based approach to abstractive summarization of
  highly redundant opinions.
\newblock In \emph{Proceedings of the 23rd International Conference on
  Computational Linguistics (Coling 2010)}, pages 340--348.

\bibitem[{Ghalandari et~al.(2020)Ghalandari, Hokamp, Pham, Glover, and
  Ifrim}]{ghalandari2020WCEP}
Demian~Gholipour Ghalandari, Chris Hokamp, Nghia~The Pham, John Glover, and
  Georgiana Ifrim. 2020.
\newblock A large-scale multi-document summarization dataset from the wikipedia
  current events portal.
\newblock \emph{arXiv preprint arXiv:2005.10070}.

\bibitem[{Gillick and Favre(2009)}]{gillick2009scalable}
Dan Gillick and Benoit Favre. 2009.
\newblock A scalable global model for summarization.
\newblock In \emph{Proceedings of the Workshop on Integer Linear Programming
  for Natural Langauge Processing}, pages 10--18. Association for Computational
  Linguistics.

\bibitem[{Goldstein et~al.(2000)Goldstein, Mittal, Carbonell, and
  Kantrowitz}]{goldstein2000multi}
Jade Goldstein, Vibhu Mittal, Jaime Carbonell, and Mark Kantrowitz. 2000.
\newblock Multi-document summarization by sentence extraction.
\newblock In \emph{Proceedings of the 2000 NAACL-ANLP Workshop on Automatic
  summarization}, pages 40--48. Association for Computational Linguistics.

\bibitem[{Gu et~al.(2020)Gu, Mao, Han, Liu, Yu, Wu, Yu, Finnie, Zhai, and
  Zukoski}]{newshead2020}
Xiaotao Gu, Yuning Mao, Jiawei Han, Jialu Liu, Hongkun Yu, You Wu, Cong Yu,
  Daniel Finnie, Jiaqi Zhai, and Nicholas Zukoski. 2020.
\newblock {Generating Representative Headlines for News Stories}.
\newblock In \emph{Proc. of the the Web Conf. 2020}.

\bibitem[{Haghighi and Vanderwende(2009)}]{haghighi2009exploring}
Aria Haghighi and Lucy Vanderwende. 2009.
\newblock Exploring content models for multi-document summarization.
\newblock In \emph{Proceedings of Human Language Technologies: The 2009 Annual
  Conference of the North American Chapter of the Association for Computational
  Linguistics}, pages 362--370. Association for Computational Linguistics.

\bibitem[{Hong and Nenkova(2014)}]{hong2014improving}
Kai Hong and Ani Nenkova. 2014.
\newblock Improving the estimation of word importance for news multi-document
  summarization.
\newblock In \emph{Proceedings of the 14th Conference of the European Chapter
  of the Association for Computational Linguistics}, pages 712--721.

\bibitem[{Lample and Conneau(2019)}]{lample2019cross}
Guillaume Lample and Alexis Conneau. 2019.
\newblock Cross-lingual language model pretraining.
\newblock \emph{Advances in Neural Information Processing Systems (NeurIPS)}.

\bibitem[{Lebanoff et~al.(2018)Lebanoff, Song, and
  Liu}]{lebanoff-etal-2018-adapting}
Logan Lebanoff, Kaiqiang Song, and Fei Liu. 2018.
\newblock \href {https://doi.org/10.18653/v1/D18-1446} {Adapting the neural
  encoder-decoder framework from single to multi-document summarization}.
\newblock In \emph{Proceedings of the 2018 Conference on Empirical Methods in
  Natural Language Processing}, pages 4131--4141, Brussels, Belgium.
  Association for Computational Linguistics.

\bibitem[{Lewis et~al.(2019)Lewis, Liu, Goyal, Ghazvininejad, Mohamed, Levy,
  Stoyanov, and Zettlemoyer}]{lewis2019bart}
Mike Lewis, Yinhan Liu, Naman Goyal, Marjan Ghazvininejad, Abdelrahman Mohamed,
  Omer Levy, Veselin Stoyanov, and Luke Zettlemoyer. 2019.
\newblock Bart: Denoising sequence-to-sequence pre-training for natural
  language generation, translation, and comprehension.
\newblock \emph{arXiv preprint arXiv:1910.13461}.

\bibitem[{Lin and Bilmes(2011)}]{lin2011class}
Hui Lin and Jeff Bilmes. 2011.
\newblock A class of submodular functions for document summarization.
\newblock In \emph{Proceedings of the 49th Annual Meeting of the Association
  for Computational Linguistics: Human Language Technologies-Volume 1}, pages
  510--520. Association for Computational Linguistics.

\bibitem[{Liu et~al.(2018)Liu, Saleh, Pot, Goodrich, Sepassi, Kaiser, and
  Shazeer}]{liu2018generating}
Peter~J. Liu, Mohammad Saleh, Etienne Pot, Ben Goodrich, Ryan Sepassi, Lukasz
  Kaiser, and Noam Shazeer. 2018.
\newblock \href {https://openreview.net/forum?id=Hyg0vbWC-} {Generating
  wikipedia by summarizing long sequences}.
\newblock In \emph{International Conference on Learning Representations}.

\bibitem[{Liu and Lapata(2019)}]{LiuL19-hierarchical}
Yang Liu and Mirella Lapata. 2019.
\newblock \href {https://doi.org/10.18653/v1/p19-1500} {Hierarchical
  transformers for multi-document summarization}.
\newblock In \emph{Proceedings of the 57th Conference of the Association for
  Computational Linguistics, {ACL} 2019, Florence, Italy, July 28- August 2,
  2019, Volume 1: Long Papers}, pages 5070--5081. Association for Computational
  Linguistics.

\bibitem[{Liu et~al.(2019)Liu, Ott, Goyal, Du, Joshi, Chen, Levy, Lewis,
  Zettlemoyer, and Stoyanov}]{liu2019roberta}
Yinhan Liu, Myle Ott, Naman Goyal, Jingfei Du, Mandar Joshi, Danqi Chen, Omer
  Levy, Mike Lewis, Luke Zettlemoyer, and Veselin Stoyanov. 2019.
\newblock Roberta: A robustly optimized bert pretraining approach.
\newblock \emph{arXiv preprint arXiv:1907.11692}.

\bibitem[{Nallapati et~al.(2016)Nallapati, Zhou, dos Santos,
  GuÌ‡l{\c{c}}ehre, and Xiang}]{nallapati-etal-2016-abstractive}
Ramesh Nallapati, Bowen Zhou, Cicero dos Santos, {\c{C}}a{\u{g}}lar
  GuÌ‡l{\c{c}}ehre, and Bing Xiang. 2016.
\newblock \href {https://doi.org/10.18653/v1/K16-1028} {Abstractive text
  summarization using sequence-to-sequence {RNN}s and beyond}.
\newblock In \emph{Proceedings of The 20th {SIGNLL} Conference on Computational
  Natural Language Learning}, pages 280--290, Berlin, Germany. Association for
  Computational Linguistics.

\bibitem[{Nayeem et~al.(2018)Nayeem, Fuad, and Chali}]{nayeem2018abstractive}
Mir~Tafseer Nayeem, Tanvir~Ahmed Fuad, and Yllias Chali. 2018.
\newblock Abstractive unsupervised multi-document summarization using
  paraphrastic sentence fusion.
\newblock In \emph{Proceedings of the 27th International Conference on
  Computational Linguistics}, pages 1191--1204.

\bibitem[{Owczarzak and Dang(2011)}]{owczarzak2011overview}
Karolina Owczarzak and Hoa~Trang Dang. 2011.
\newblock Overview of the tac 2011 summarization track: Guided task and aesop
  task.
\newblock In \emph{Proceedings of the Text Analysis Conference (TAC 2011),
  Gaithersburg, Maryland, USA, November}.

\bibitem[{Paul and James(2004)}]{paul2004introduction}
Over Paul and Yen James. 2004.
\newblock An introduction to duc-2004.
\newblock In \emph{Proceedings of the 4th Document Understanding Conference
  (DUC 2004)}.

\bibitem[{Radev et~al.(2004)Radev, Jing, Sty{\'s}, and Tam}]{radev2004centroid}
Dragomir~R Radev, Hongyan Jing, Ma{\l}gorzata Sty{\'s}, and Daniel Tam. 2004.
\newblock Centroid-based summarization of multiple documents.
\newblock \emph{Information Processing \& Management}, 40(6):919--938.

\bibitem[{Raffel et~al.(2019)Raffel, Shazeer, Roberts, Lee, Narang, Matena,
  Zhou, Li, and Liu}]{2019t5}
Colin Raffel, Noam Shazeer, Adam Roberts, Katherine Lee, Sharan Narang, Michael
  Matena, Yanqi Zhou, Wei Li, and Peter~J. Liu. 2019.
\newblock \href {http://arxiv.org/abs/1910.10683} {Exploring the limits of
  transfer learning with a unified text-to-text transformer}.
\newblock \emph{arXiv e-prints}.

\bibitem[{Wolf et~al.(2019)Wolf, Debut, Sanh, Chaumond, Delangue, Moi, Cistac,
  Rault, Louf, Funtowicz, and Brew}]{Wolf2019HuggingFacesTS}
Thomas Wolf, Lysandre Debut, Victor Sanh, Julien Chaumond, Clement Delangue,
  Anthony Moi, Pierric Cistac, Tim Rault, R'emi Louf, Morgan Funtowicz, and
  Jamie Brew. 2019.
\newblock Huggingface's transformers: State-of-the-art natural language
  processing.
\newblock \emph{ArXiv}, abs/1910.03771.

\bibitem[{Yasunaga et~al.(2017)Yasunaga, Zhang, Meelu, Pareek, Srinivasan, and
  Radev}]{yasunaga2017graph}
Michihiro Yasunaga, Rui Zhang, Kshitijh Meelu, Ayush Pareek, Krishnan
  Srinivasan, and Dragomir Radev. 2017.
\newblock Graph-based neural multi-document summarization.
\newblock In \emph{Proceedings of the 21st Conference on Computational Natural
  Language Learning (CoNLL 2017)}, pages 452--462.

\bibitem[{Zopf(2018)}]{zopf2018auto}
Markus Zopf. 2018.
\newblock Auto-hmds: Automatic construction of a large heterogeneous
  multilingual multi-document summarization corpus.
\newblock In \emph{Proceedings of the Eleventh International Conference on
  Language Resources and Evaluation (LREC 2018)}.

\bibitem[{Zopf et~al.(2018)Zopf, Menc{\'\i}a, and
  F{\"u}rnkranz}]{zopf2018scores}
Markus Zopf, Eneldo~Loza Menc{\'\i}a, and Johannes F{\"u}rnkranz. 2018.
\newblock Which scores to predict in sentence regression for text
  summarization?
\newblock In \emph{Proceedings of the 2018 Conference of the North American
  Chapter of the Association for Computational Linguistics: Human Language
  Technologies, Volume 1 (Long Papers)}, pages 1782--1791.

\end{thebibliography}
\bibliographystyle{acl_natbib}

% \clearpage
% \appendix
% \section{Appendix}
% \input{appendix}

\end{document}